\documentclass[letterpaper,10pt,conference]{ieeeconf}  

\IEEEoverridecommandlockouts

\usepackage{microtype}
\usepackage{caption}
\usepackage{subcaption}
\usepackage{booktabs}
\usepackage{balance}
\usepackage{xcolor}
\usepackage{tabularx}
\usepackage{graphicx,dblfloatfix} 
\usepackage{amsmath}
\usepackage{amssymb}
\usepackage{listings}
\usepackage{textcomp}
\usepackage{url}
\usepackage{multirow}
\usepackage{algorithm2e}
\usepackage{todonotes}
\usepackage{hyperref}
\usepackage{cleveref}
\usepackage{wrapfig}
\crefformat{footnote}{#2\footnotemark[#1]#3}
\usepackage{ragged2e} 
\usepackage[subnum]{cases}
\newcolumntype{Y}{>{\RaggedRight\arraybackslash}X} 
\usepackage[noend]{algpseudocode}

\usepackage{tikz}
\usepackage{tkz-euclide}
\usetikzlibrary{math}
\usepackage{pgfplots}
\usetikzlibrary{arrows}

\usepackage{float}

\usepackage{multicol, blindtext}

\usepackage{multirow}
\usepackage{adjustbox}
\usepackage{booktabs}
\usepackage{footnote}

\usepackage{bm}

\newcommand{\comment}[1]{}

\newcommand{\alloc}[1]{}

\newcommand{\emp}{\mathrm{emp}}

\newcommand{\occ}{\mathrm{occ}}

\title{OHM: GPU Based Occupancy Map Generation}
\author{Kazys Stepanas,$^{1}$ Jason Williams,$^{1}$ Emili Hern\'{a}ndez,$^{2}$ Fabio Ruetz,$^{1,3}$ Thomas Hines$^{1}$    
\thanks{$^{1}$ K. Stepanas, J. Williams, F. Ruetz and T. Hines are with the Robotics and Autonomous Systems Group (RASG), CSIRO, Pullenvale, QLD 4069, Australia. All correspondence should be addressed to {\tt\small kazys.stepanas@csiro.au}}
\thanks{$^{2}$ E. Hernandez is with Emesent, Milton, QLD 4064, Australia and was with CSIRO RASG at the time of this work.}
\thanks{$^{3}$ F. Ruetz is also with QUT Centre for Robotics, Brisbane QLD 4000.}
\thanks{This research was developed with funding from the Defense Advanced Research Projects Agency (DARPA). The views, opinions and/or findings expressed are those of the author and should not be interpreted as representing the official views or policies of the Department of Defense or the U.S. Government.}
}

\begin{document}

\maketitle
\thispagestyle{empty}
\pagestyle{empty}

\begin{abstract}
Occupancy grid maps (OGMs) are fundamental to most systems for autonomous robotic navigation. However, CPU-based implementations struggle to keep up with data rates from modern 3D lidar sensors, and provide little capacity for modern extensions which maintain richer voxel representations. This paper presents OHM, our open source, GPU-based OGM framework. We show how the algorithms can be mapped to GPU resources, resolving difficulties with contention to obtain a successful implementation. The implementation supports many modern OGM algorithms including NDT-OM, NDT-TM, decay-rate and TSDF. A thorough performance evaluation is presented based on tracked and quadruped UGV platforms and UAVs, and data sets from both outdoor and subterranean environments. The results demonstrate excellent performance improvements both offline, and for online processing in embedded platforms. Finally, we describe how OHM was a key enabler for the UGV navigation solution for our entry in the DARPA Subterranean Challenge, which placed second at the Final Event.  
\end{abstract}

\section{Introduction}
\label{sec:Introduction}
\bstctlcite{IEEEexample:BSTcontrol}Since their inception in \cite{elfes89}, occupancy grid maps (OGMs) have played a fundamental role in integrating raw sensor data into a form suitable for exploitation in autonomous navigation across a wide range of applications. Advances in sensing have yielded progressively higher sensor resolutions for modern lidar sensor data rates (Table~\ref{tab:LidarSamplesPerSecond}), and enabled navigation in complex 3D environments. However, integration of this highly informative data is largely impractical in modern embedded CPUs.

Research over the past decade has yielded extensions of OGMs which retain richer information within each voxel. These include Normal Distributions Transform (NDT) Occupancy Maps (NDT-OM) \cite{Saarinen2013}, NDT-Traversability Maps (NDT-TM) \cite{AhtSto17}, Decay rate maps \cite{SchLuf17}, and Truncated Sign Distance Functions (TSDF) \cite{Oleynikova2017}. These representations maintain richer representations within each voxel than the traditional probability of occupancy, providing sub-voxel information that can enhance navigation performance; e.g., for estimating traversability, or indicating the distance to the closest obstacle. However, the computational cost of maintaining many of these representations limits their applicability in online systems.

GPU resources on modern platforms are often under-utilised, due to the difficulty of mapping algorithms onto the massive parallelism that they expose. In this paper, we describe our open source Occupancy Homogeneous Mapping (OHM) system, examining the design and approach for addressing contention, the key problem in such a system. To our knowledge, this challenging development has not been documented elsewhere in the literature. OHM is designed to be extensible; we demonstrate this by incorporating support for the modern variants above, specifically, NDT-OM, NDT-TM, decay rate maps and TSDFs. OHM incorporates GPU implementations in CUDA (supporting NVIDIA GPUs) and OpenCL (tested in our systems in integrated Intel GPUs), alongside reference CPU implementations that aid in development. 

The practical implications of OHM are significant. We utilised its NDT-OM representation in all of the ground platforms fielded by Team CSIRO Data61 in the DARPA Subterranean Challenge (SubT) \cite{subt}. On the Boston Dynamics Spot platform, OHM enabled the both lidar and integrated depth cameras to be integrated using the NUC's integrated Intel GPU, which was otherwise unused, freeing up CPU resources and permitting complex navigation including autonomous ascent and descent of stairs. We believe that the relatively low weight of our payload (enabled by OHM's use of the GPU) contributed significantly to our team's strong performance in the SubT Final Event.

\begin{table}
\centering
\caption{Approximate, maximum samples per second generated by some multi-beam lidar systems. Values are in million samples/second given for single return mode only.}
\begin{tabular}{lc}
\toprule
Lidar & $10^6$ samples/s \\
\midrule
Ouster OS 32 beam & 0.66 \\ 
Ouster OS 64 beam & 1.3 \\
Ouster OS 128 beam & 2.6 \\
Velodyne VLP-16 & 0.3 \\
Velodyne VLP-32c & 0.6 \\
Velodyne VLS-128 & 2.3 \\
\bottomrule
\end{tabular}
\label{tab:LidarSamplesPerSecond}
\vspace{-0.5cm}
\end{table}

\subsection{Contributions}
\label{ss:Contributions}
The contributions of this work are:
\begin{itemize}
\item We describe the design of OHM, focusing on the memory model and handling of contention, which enables exploitation of GPU parallelism for OGMs
\item We outline modern extensions of occupancy mapping, NDT-OM, NDT-TM, decay rate maps and TSDFs, and describe the steps taken to incorporate them into OHM; to our knowledge neither a CPU nor a GPU implementation of NDT-OM, NDT-TM or decay rate maps is publicly available at the current time
\item We provide an extensive computational evaluation, comparing OHM to state-of-the-art alternatives, Octomap \cite{hornung13auro} and Voxblox \cite{Oleynikova2017} using real-world captured data both offline on a performance laptop, and online on the embedded platforms utilised in our ground robots
\item The data sets used for comparison were chosen to include diverse environments including outdoor, mixed outdoor-indoor data, and underground data collected during SubT, from quadruped and tracked UGV platforms as well as UAVs; these will be released to enable reproduction of results and comparison to new approaches
\item We describe the employment of OHM in our successful entry in SubT; we believe that OHM was a key enabler to our successful second place result
\end{itemize}
OHM is open source \cite{github:ohm}, incorporating CUDA and OpenCL GPU implementations, alongside a reference CPU implementation which is useful in development.

\section{Related work}

We begin this section by outlining a cross-section of the vast array of OGM applications. Subsequently, we survey the representations and updates in OGMs, and their modern extensions implemented in OHM. Strategies that have been employed for efficient implementation are also described.

\subsection{Applications}
\label{ss:Applications}

Since the seminal work of a ground vehicle navigating in an office environment~\cite{elfes89}, OGMs have been widely used in multiple applications and domains, becoming the \emph{de-facto} local (or short term memory) navigation solution in robotics. 

On ground vehicles they have been successfully applied in multiple robotic navigation frameworks in a wide range of environments, from fully structured offices~\cite{Marder-Eppstein2010} to heavily unstructured rainforests~\cite{yue2020}. In those scenarios, 2D and 3D OGMs are commonly used for motion planning and obstacle avoidance~\cite{tanzmeister2014} as well as waypoint navigation and exploration~\cite{hines2021} in both static and changing environments~\cite{meyer-delius2012}. 

OGMs are heavily used in the aerial domain on multi-rotor applications that require of safe navigation in challenging or cluttered environments, such as underground mines~\cite{hovermap, akbari2021}, vineyards~\cite{santos2021} as well as forest environments~\cite{hovermap}.

In marine robotics it is common to make use OGMs on surface vessels in complex environments such as harbours to identify docking-based structures~\cite{pereira2021}. Underwater robots make use of OGMs for map building~\cite{Hernandez2009} and motion planning purposes~\cite{jd_hernandez2015, Hernandez2019, Vidal2020}, mainly in feature rich environments such as such as marinas, ports and water-breaks.

OGMs play an important role in industrial robotic motion planning applications as depicted in the supported robotic platforms webpage of MoveIt~\cite{moveit}, especially in cooperative scenarios in which operators share the same space~\cite{Diftler2011}. They are also used extensively in the self-driving car industry, since they can deal with a wide variety of environments under diverse conditions of illumination and traffic~\cite{wall_mutz2021}, addressing problems such as road boundary detection~\cite{Homm2010}. In this domain performance is a critical factor and methods tend to take full advantage of GPU implementations~\cite{Fickenscher2018}.

The OHM implementation presented in this paper has been used widely in UGV navigation applications on wheeled, tracked, quadruped and hexapod platforms~\cite{HudTal21}, as well as UAV navigation and planning and offline point cloud processing and filtering.

\subsection{Occupancy grid maps}
\label{ss:OccMap}
Since being introduced in~\cite{elfes89}, OGMs have become pervasive in mobile robotics, due to their ability to leverage probabilistic sensor models to integrate information from multiple sensors and times. The framework models occupancy as a Markov random field of zero-th order, such that cell states remain independent. The Bayesian update of occupancy probability is:
\begin{equation}\label{eq:BasicProbUpdate}
P(C_i|z^t) = \frac{
P(z_t|C_i) P(C_i|z^{t-1})
}{
\sum_{c\in\{\emp,\occ\}}
P(z_t|C_i=c) P(C_i=c|z^{t-1}) 
}
\end{equation}
where, $C_i\in\{\emp,\occ\}$ denotes the state of cell $i$ as empty or occupied, $z_t$ is the sensor reading at time $t$, and $z^t=(z_0,\dots,z_t)$ denotes the history of sensor readings. When a return falls within a cell, $P(z_t|C_i=\occ)>P(z_t|C_i=\emp)$, reinforcing the likelihood of the cell being occupied, while an observation that passes through a cell thus obtaining evidence that the cell is free space has $P(z_t|C_i=\occ)<P(z_t|C_i=\emp)$. It is common to implement these updates in log-odds form to avoid numerical instabilities at the probability range extremes, representing updates as
\begin{align}
l(C_i=\occ|z^t) &= l(C_i=\occ|z^{t-1}) + \delta l_i \label{eq:BasicLogOdds} \\
\delta l_i &= \log \frac{P(z_t|C_i=\occ)}{P(z_t|C_i=\emp)}
\end{align}
It can be easily shown that \eqref{eq:BasicProbUpdate} and \eqref{eq:BasicLogOdds} are equivalent through the invertible transformation $l(C_i=\occ|z^t)=\frac{P(C_i=\occ|z^t)}{P(C_i=\emp|z^t)}$. We subsequently refer to the latest occupancy likelihood for voxel $i$ as $l_i\triangleq l(C_i=\occ|z^t)$.

It is common to clamp these values, such that they are prevented from becoming larger or smaller than certain thresholds. This provides an ability to handle changes in the environment; without saturation, a large amount of evidence accrued in one state would take a long time to counteract in order to infer that the state has changed.

\subsection{3D Occupancy Grid Maps}
\label{sec:3DOGM}
Mathematically, 2D OGMs extend trivially to 3D, applying the same update expressions along rays passing through a 3D grid of voxels. While 3D OGMs were anticipated in earlier work, they did have not become practical for online implementation until recent years. One of the most popular 3D OGM approaches at the current time is Octomap~\cite{hornung13auro}. The key advance of this approach is that it provides memory-efficient representation of OGMs. Specifically, an octree is formed where leaves are voxels as in a regular 3D OGM, with probability of occupancy updated as in \eqref{eq:BasicLogOdds}. Each leaf can be in one of three states, \textit{occupied} (clamped), \textit{free} (clamped) and \textit{uncertain} (not yet clamped). The occupancy probability is only required for the \textit{uncertain} state, so whenever all descendants of node in the tree are in the same occupied or free state, the node can act as a sufficient statistic for all of its descendants.
However, with a lidar sensor, there is usually not a contiguous solid angle observed, and the situation often arises where some clamped voxels are intermixed with unknown and uncertain voxels (particularly at longer ranges), preventing wide use of branch collapse. %
Additionally, the tree structure increases in the computation required to query and update individual voxels; e.g., in a map with $K$ voxels, the cost of querying a particular voxel is $O(\log K)$, whereas voxels can be addressed through $O(1)$ operations in a homogeneous occupancy map.

\subsection{Normal Distribution Transform OM and TM}
\label{ss:NDT}
Classical OGMs often suffer from erosion on surfaces, where returns may fall in adjacent voxels depending on perspective and random noise. Methods using NDT-OM~\cite{Saarinen2013} address this difficulty by retaining information on the location of returns within voxels. This also permits modelling of small objects within a voxel (e.g., a narrow wire or pole) which yield a mixture of returns and pass-through rays.

Using NDT-OM each voxel is endowed with a mean vector, and covariance matrix, in addition to the occupancy likelihood. NDT-OM uses the distribution of points within a voxel to minimise the reduction in occupancy probability when a ray does not intersect the occupied portion of the voxel, thus reducing erosion. The update of the mean and covariance uses efficient recursive calculations.

NDT-TM~\cite{AhtSto17} extends NDT-OM by additionally calculating the permeability (i.e., transparency) of voxels, along with the distribution of intensity in each voxel. The permeability estimate for voxel $i$ is calculated from the number of hits $H_i$ and number of misses $M_i$ as $\hat{p}_i = \frac{H_i}{H_i+M_i}$; $H_i$ and $M_i$ are updated with additional checks on the weighting quantities used in NDT-OM to verify that the ray is close to the data in the voxel. NDT-TM also maintains a 1D Gaussian of intensity, and NDT-TM utilises these all these quantities as features in a Support Vector Machine (SVM) classifier, which distinguishes between traversable and non-traversable terrain. 

In the OHM implementation of both NDT-OM and NDT-TM clamping is applied for the occupancy as described above, and sample data is cleared when the voxel occupancy likelihood falls below a threshold. Again, this provides a basic capability for addressing transient objects.

\subsection{Decay Rate Map}
\label{ss:DecayRate}
The decay rate map introduced in~\cite{SchLuf17} provides an alternate formulation that utilises additional information beyond the traditional binary occupancy. Reflections are modelled as occurring at a constant rate $\lambda$ within a voxel. Under this model, it can be shown that the reflection rate within a voxel $i$ can be estimated as $\lambda_i = \frac{H_i}{\sum_{j\in\mathcal{J}} d_i(j)}$ where $H_i$ is the total number of returns falling in voxel $i$, and $d_i(j)$ is the distance that the $j$-th ray travels through voxel $i$. As well as providing an analytic derivation, the ray-path information that is captured is shown to improve performance when used as a part of a SLAM framework.

\subsection{Sign-distance functions}
\label{ss:SDF}
Another popular group of 3D map representation for local navigation use Signed-Distance Functions (SDFs)~\cite{curless1996volumetric}. Commonly these maps are discretised using 3D voxel representations, with each voxel containing a SDF value and weight. The Euclidean SDF (ESDF) retains the distance of the closest surface for each voxel, resulting in an implicit representation of the geometry of the environment. The sign is positive for voxels that lie outside of a surface, zero on the surface boundary and negative if inside.
Calculating these distances with sequential sensor readings can be computationally involved and led to the development of the Truncated SDF (TSDF)~\cite{newcombe2011}. TSDFs use a projected distance function, based on current range measurements. TSDF updates the voxels, up to a truncation distance, in front and behind current measurement along the ray given by the sensor origin and measurement point. Voxblox~\cite{Oleynikova2017} provides a CPU implementation designed to work with lidar and RGBD, which dynamically grows a dense map with a sequential ESDF-map generation suitable for planning. State-of-the-art performance is showcased through real-world demonstrations on autonomous UAVs in unstructured environments. 

While SDFs have benefits over classical representations in their ability to encode sub-voxel resolution, they inherently can represent one surface per voxel, and viewing a surface from multiple angles will erode it. This issue is compounded if there are multiple thin, pole-like objects or wire meshes within. Hence, natural areas with thin grass blades, caves with stalactites or trees are challenging environments for this representation as it does not retain statistics over a longer period. This is problematic if the representation is used for navigation purposes.

\section{GPU Based Approach}

As new lidars are being introduced into the market offering higher specifications in both number of beams and increased range, the CPU-based OGM approaches struggle to keep pace with the incoming data stream. The prevalence of multiple CPU cores in embedded platforms allows multi-core, thread-safe adaptations of the voxel algorithms above, increasing the data rate capabilities. However, the addition of each new CPU core shows diminishing returns in processing bandwidth as shown in Table~\ref{tab:TsdfThreads}. In addition to multiple CPU cores, embedded platforms often have GPU resources available, which are frequently under-utilised or left idle. By leveraging GPU resources we show that the processing bandwidth may be dramatically increased, while leaving CPU resources available for other algorithms less suited to GPU conversion.

\begin{table}
\vspace*{4pt}
\centering
  \caption{Diminishing returns of additional threads when processing the SubT finals platform data set using the Voxblox TSDF fast algorithm.}
  \begin{tabular}{ c c c c }
    \toprule
    Threads & $10^6$ rays/s processed & Rays/s delta & \% delta \\
    \midrule
    1 & 1.082 & N/A & N/A \\
    2 & 1.504 & 0.421 & 39 \\
    4 & 2.039 & 0.535 & 36 \\
    6 & 2.300 & 0.262 & 13 \\
    \bottomrule
  \end{tabular}
  \label{tab:TsdfThreads}
  \vspace{-0.5cm}
\end{table}

GPU implementation requires careful examination to map both the memory model and algorithm structure to the available parallelism. Choosing an appropriate memory layout is critical to enable GPU based occupancy map generation. While octrees are an efficient CPU based data structure and static octrees may be accessed in multi-threaded scenarios, dynamic octree update does not scale well to multiple CPU threads. They are even less suited to GPU based update where fixed memory layout and well-behaved branching patterns yield the highest gains. We begin this section by defining a memory layout suitable for use with GPU algorithms then define the GPU algorithms for integrating ray samples, and their extensions for accommodating modern OGM variants.

\subsection{Memory Layout}
\label{ss:MemoryLayout}
In order to support efficient data transfer and GPU based manipulation, the proposed approach divides space into discrete voxel regions. Regions are stored in a hash map, permitting dynamic growth of the mapping area; a similar approach is utilised in~\cite{Oleynikova2017}. Each region is a dense 3D grid of cubic, fixed size voxels allocated as a contiguous memory block. The contiguous allocation ensures efficient transfer between GPU and host memory, as well as defining clear access patterns for GPU code. Voxels are addressed via a region key (an index into the region hash map), and a voxel index within the region.

A GPU caching algorithm is used to manage data transfer between host and device memory. This imposes some restrictions on the maximum spatial dimensions of a single ray set integrated into the map based on the cache size. In practice this restriction is not significant with current lidar range limits and modern GPU memory sizes\footnote{See examples in documentation in \cite{github:ohm}.}. 
Offline processing of large, global maps poses a challenge due to (host) memory requirements. Since input data samples are expected to appear within a local spatial volume which moves slowly over time, we add a background process which compresses regions which have not been recently accessed.

As discussed in Section~\ref{ss:Layers}, a key design goal of OHM is extensibility, i.e., being able to support flexible voxel data. This is achieved by supporting multiple contiguous layers for each voxel region. For a simple occupancy map, each voxel has a 32\,bit floating point occupancy likelihood; in other cases, e.g., NDT variants, it may include layers for voxel means and covariances.

\subsection{Ray Integration}
\label{ss:RayIntegration}

GPU implementation requires careful consideration of the source of parallelism being exploited. For occupancy mapping, parallelism over voxels is conceptually attractive, especially since it avoids contention and allows use of fast local memory. However, for lidar this turns out to be ineffective since the vast majority of voxels are not visited in a given ray set (especially at longer ranges), resulting in poor workload efficiency.

The successful approach has been to utilise parallelism over rays, with one GPU thread per ray. This approach creates the potential for conflict where multiple GPU threads may attempt an update of the same target voxel, creating thread contention. This conflict is resolved using atomic Compare And Swap (CAS) semantics via OpenCL~\cite{webpage:opencl} and CUDA~\cite{webpage:cuda} atomic instructions. This technique imposes some inefficiencies, most notably the requirement that all GPU threads operate on global memory rather than the more efficient local memory. Nevertheless, the final performance benefits result in a significant improvement in execution time. This approach works for hit and miss updates for occupancy, mean, decay rate and TSDF voxels where the per-voxel data is 32 or 64\,bits which aligns with supported 32 and 64\,bit CAS operations. The method used to update the larger covariance matrix stored in NDT is described in Section~\ref{ss:Layers} and it is only required for updating voxels with samples falling within them, i.e., the observed end-point of the ray.

The line walking algorithm was adapted to three dimensions from~\cite{Amanatides1987AFV}.\footnote{Detailed pseudocode can be found in \cite{github:ohm}.} This algorithm steps a single voxel on each iteration selecting the a step direction based on the largest remaining distance to cover on each axis. When updating a voxel, several attempts are made to check for CAS failure due to contention. While this algorithm performs significantly better than a CPU implementation (as shown in Section~\ref{sec:Results}), it has some features that are not ideal for GPU execution. Firstly, the number of iterations of the while loop is not known before execution and is dependent on the ray length. This means that the execution time of a work group (OpenCL) or warp (CUDA) is determined by the longest ray which may significantly differ from the shorter rays. This is mitigated by breaking long rays into several segments, where all but the last have no sample and are used only for ``miss'' updates. In this way, each ray \emph{segment} can be handled by a different thread, and the irregularity of the work in different threads is bounded. Secondly, the algorithm uses global GPU memory in order correctly resolve the atomic operations. This would appear to be unavoidable in implementing the OGM algorithm.

Before executing the GPU algorithm, work must be done in CPU to manage the GPU cache. The same line walking algorithm~\cite{Amanatides1987AFV} is used to walk the \emph{regions} as opposed to \emph{voxels} in CPU. Regions are created and/or uploaded to the GPU cache as required, incurring a low CPU processing cost due to very coarse region resolution (3.2\,m in our experiments).

\subsection{Flexible Occupancy Layers}
\label{ss:Layers}
In order to meet the design goal of extensibility, each region can possess a number of layers, storing additional data. In addition to the occupancy value itself, common extensions are the voxel mean layer, which adds a 32\,bit discretised mean position\footnote{Packing 10\,bit subvoxel mean of points in $x$, $y$ and $z$ into a 32\,bit integer} and a 32\,bit sample count layer storing the number of samples which contribute to the mean.

As described in Section~\ref{ss:NDT}, NDT-OM and NDT-TM require additional quantities including the covariance matrix of points in the voxel. This is represented as a triangular square root matrix, and maintained using the numerically stable updates in~\cite{Maybeck1979}. While occupancy and mean values can be updated using CAS semantics, a more sophisticated method is required for these larger data structures. Noting that the covariance is only updated for voxels containing samples (and that the ``miss'' update remains a simple occupancy probability adjustment), this complexity is handled by splitting the update into two phases, the first of which processes ``miss'' events, and the second of which incorporates new sample points in the corresponding voxels. 
During the second phase, the mean and covariance are updated by a single GPU thread for each voxel that requires updating, i.e., a GPU thread calculates a new mean and covariance based on all samples falling in that voxel from the input set. Additional quantities for NDT-TM are handled similarly.

\section{Results}
\label{sec:Results}
We demonstrate the operation of OHM through real-time and offline processing of data sets on a range of computational platforms, based on data recorded from lidars mounted on legged, tracked and drone platforms, operating in environments ranging from narrow caves to open outdoor areas. Brief descriptions of data sets can be found in Figures~\ref{fig:ResCaveSnapshots} and \ref{fig:ResNonCaveSnapshots}. Each data set has a duration of five minutes. In each case, we compare results to state-of-the-art alternatives, specifically Voxblox~\cite{Oleynikova2017} and Octomap~\cite{hornung13auro} with OpenMP disabled since it was found it have no beneficial effect during our empirical tests. All data was collected with a Velodyne VLP-16 Puck lidar (either flat or spinning on a 30$^\circ$ inclined pedestal), with dual return mode enabled. Sensor trajectory information is provided from the Wildcat SLAM pipeline described in~\cite{HudTal21}. This sensor produces around 300,000\,rays/s.

\subsection{Offline Results}
Comparison through offline processing allows an evaluation of the maximum rate of rays that can be processed through each pipeline. This permits extrapolation of the results to other configurations involving additional beams and/or multiple sensors. All offline results were processed on a Dell Precision 7530 laptop (manufactured in 2018) with an Intel i9-8950HK processor and 32\,GB RAM, utilising the integrated Intel UHD Graphics 630 through OpenCL 2.0, and the NVIDIA Quadro P3200 GPU through CUDA 10.2, running on GNU/Linux Ubuntu 18.04. 

\begin{figure*}
\centering
\begin{subfigure}[b]{0.45\textwidth}
\centering
\includegraphics[width=\textwidth]{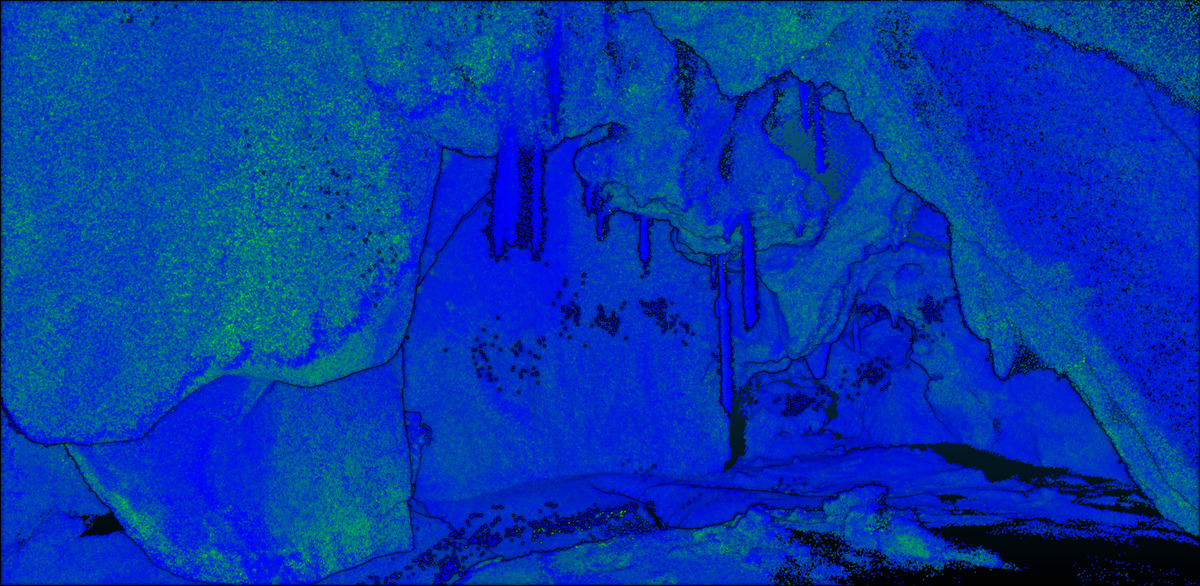}
\caption{Point cloud}
\label{fig:ResCaveCloud}
\end{subfigure}
\hfill
\begin{subfigure}[b]{0.45\textwidth}
\centering
\includegraphics[width=\textwidth]{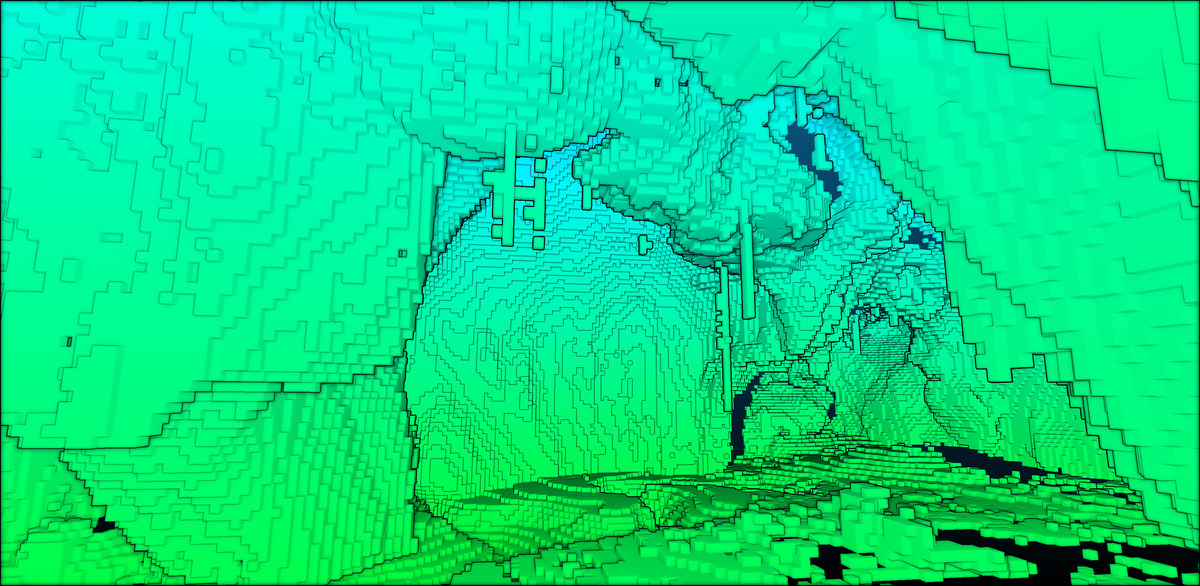}
\caption{Occupancy map}
\label{fig:ResCaveOcc}
\end{subfigure}
\hfill
\begin{subfigure}[b]{0.45\textwidth}
	\centering
	\includegraphics[width=\textwidth]{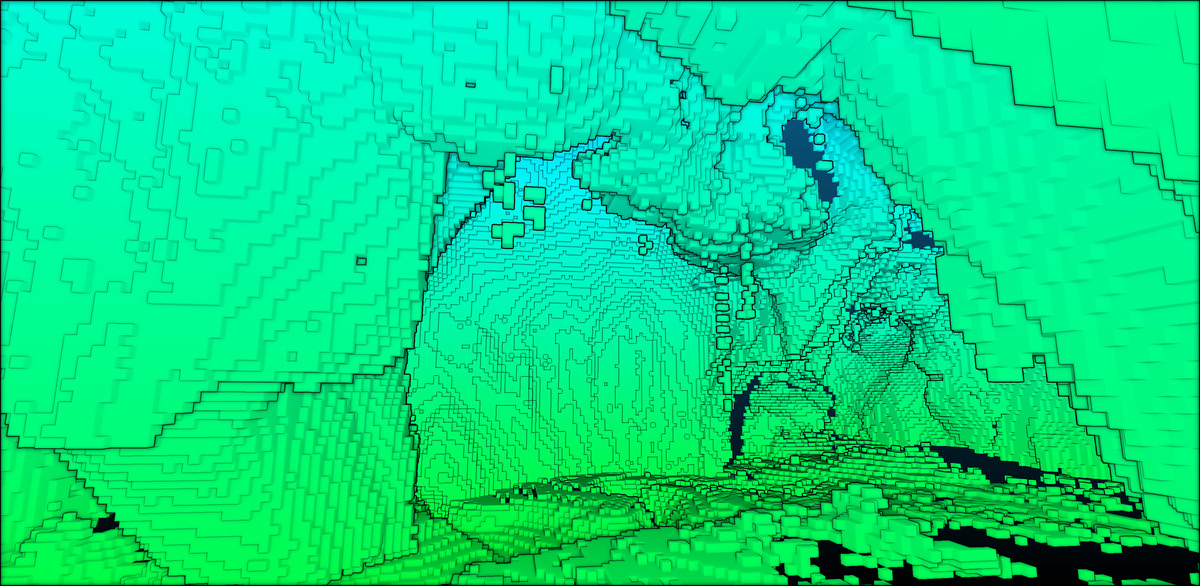}
	\caption{TSDF}
	\label{fig:ResCaveTSDF}
\end{subfigure}
\hfill
\begin{subfigure}[b]{0.45\textwidth}
	\centering
	\includegraphics[width=\textwidth]{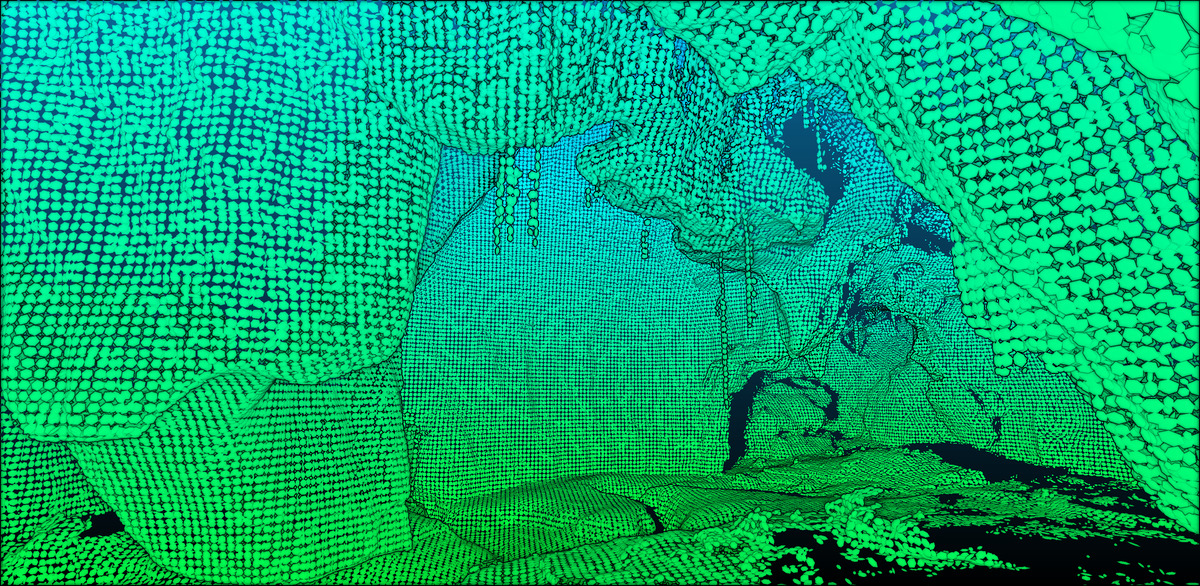}
	\caption{NDT-OM}
	\label{fig:ResCaveNDT}
\end{subfigure}
\caption{Illustration of cave data set and results of different OHM-based algorithms. The data set was captured from a BIA5 ATR tracked platform in the Carpenteria Cave systems, a naturally formed limestone cave in Chillagoe, Queensland, during autonomous navigation testing performed for SubT. For occupancy map, cubes are positioned using the calculated voxel mean, while TSDF shows voxel positions rather than the reconstructed ESDF surface.}
\label{fig:ResCaveSnapshots}
\end{figure*}

\begin{figure*}
\centering
\begin{subfigure}[b]{0.45\textwidth}
\centering
\includegraphics[width=\textwidth]{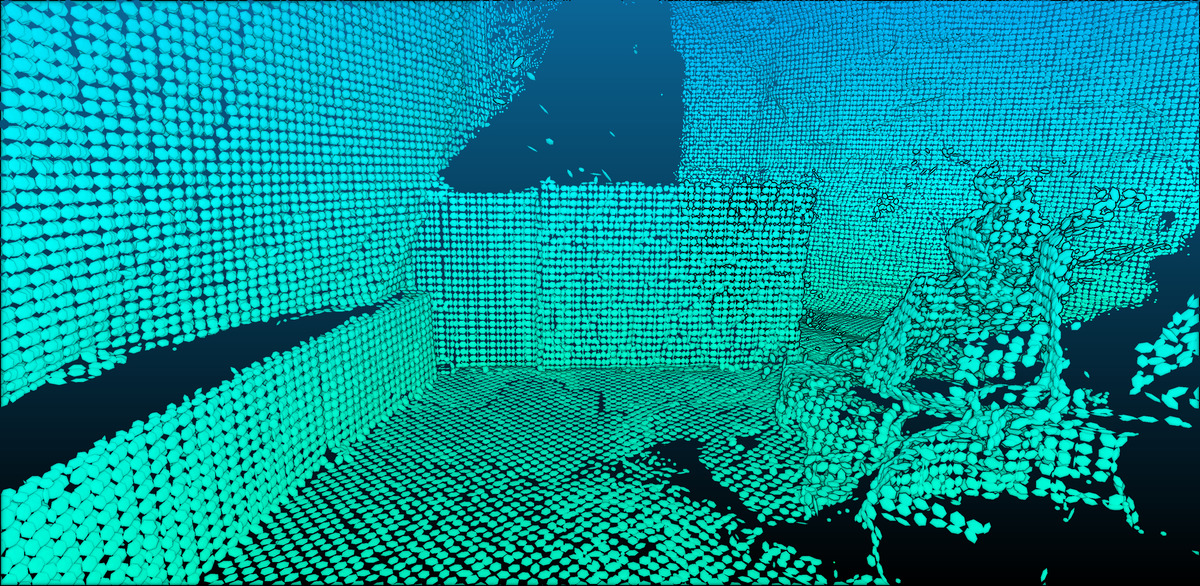}
\caption{Dust}
\label{fig:ResDust}
\end{subfigure}
\hfill
\begin{subfigure}[b]{0.45\textwidth}
	\centering
	\includegraphics[width=\textwidth]{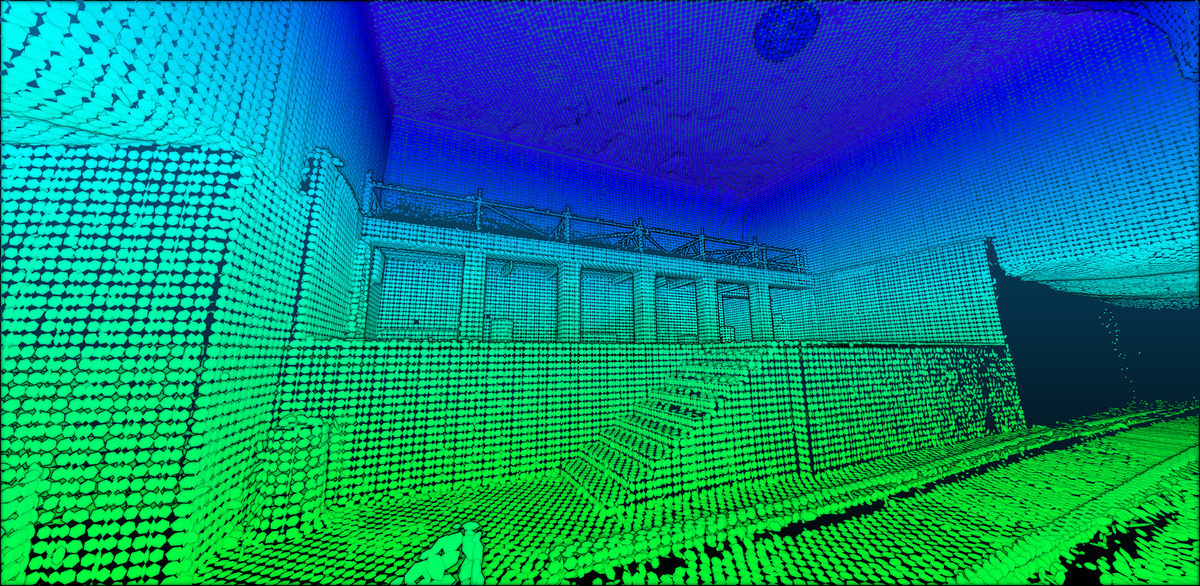}
	\caption{Railway platform}
	\label{fig:ResPlatform}
\end{subfigure}
\hfill
\begin{subfigure}[b]{0.45\textwidth}
\centering
\includegraphics[width=\textwidth]{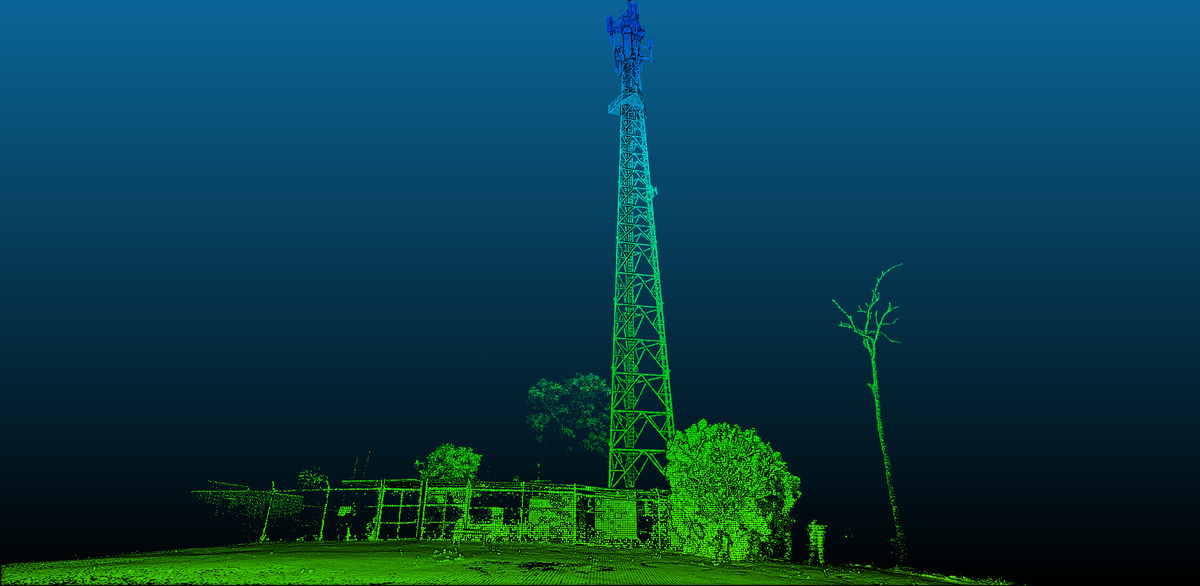}
\caption{Hovermap}
\label{fig:ResHovermap}
\end{subfigure}
\hfill
\begin{subfigure}[b]{0.45\textwidth}
	\centering
	\includegraphics[width=\textwidth]{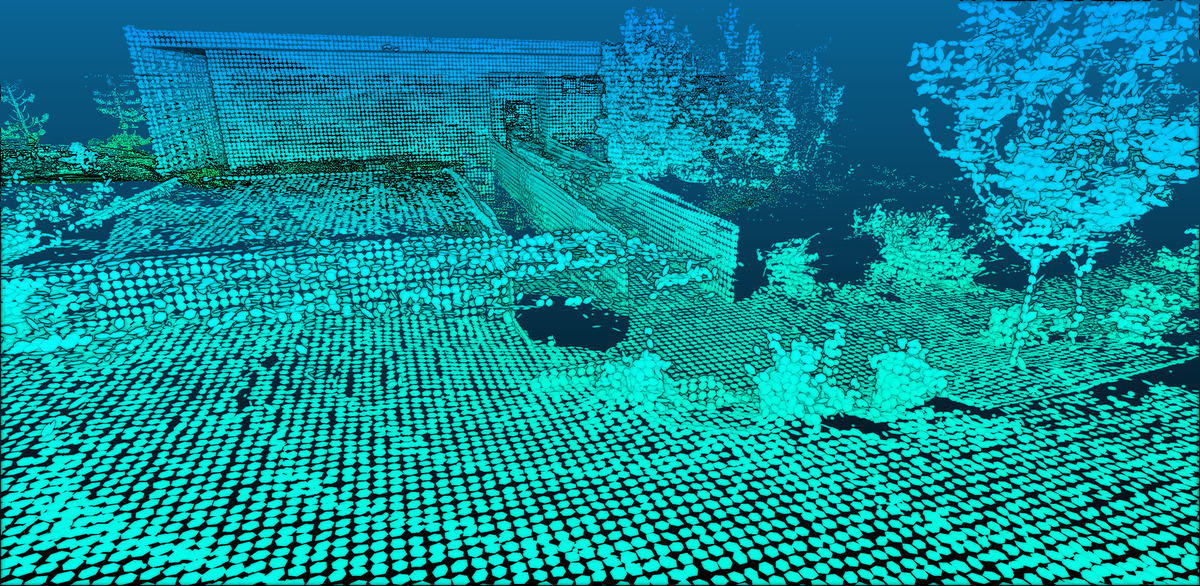}
	\caption{Flatpack}
	\label{fig:ResFlatpack}
\end{subfigure}
\caption{Illustration of data sets, through ellipsoids calculated as a part of NDT-OM representation. Dust data set (a) was captured during the SubT Final Event, from a BIA5 ATR tracked robot with an Emesent drone flying nearby, resulting in significant dust concentrations; (b) was captured during the SubT Final Event using a Boston Dynamics Spot quadruped in a portion of the course involving a railway platform and stairs down to the railway; (c) was captured using an drone carrying an Emesent Hovermap payload; and (d) was captured from a Spot quadruped with a statically mounted lidar following a course which includes both indoor and outdoor segments. Data sets other than (d) utilise an lidar mounted on an inclined, rotating pedestal.}
\label{fig:ResNonCaveSnapshots}
\end{figure*}

The first (cave) environment is illustrated through point cloud, occupancy voxel grid and NDT in Figure~\ref{fig:ResCaveSnapshots}, showing the environment itself and the data captured through the different approaches. It can be seen that OGMs and NDT-OM exhibit reduced erosion of stalactites than TSDF. Remaining data sets are illustrated through the NDT illustrations in Figure~\ref{fig:ResNonCaveSnapshots}. Animations of the online processing are provided in the video accompanying the paper.

\begin{table}
\vspace*{4pt}
\centering
\caption{Rays/s in offline processed using different methods based on various data sets using. All GPU algorithms use reverse ray tracing. Values are shown in 10\textsuperscript{6} rays/s.}
\begin{tabular}{lrrrr}
\toprule 
\textbf{Method} & \textbf{Occupancy} & \textbf{NDT-OM} & \textbf{NDT-TM} & \textbf{TSDF} \\
\midrule
\multicolumn{5}{l}{\textit{Cave data set}} \\
\midrule
OHM CPU
& 0.230 & 0.164 & 0.144 & 0.164 \\ 
OHM CUDA
& 2.721 & 1.899 & 1.800 & 2.540 \\ 
OHM OpenCL
& 1.424 & 0.855 & 0.729 & 1.059 \\ 
Octomap & 0.174 &  &  &  \\ 
Voxblox 1-thr & 0.170 &  &  &
0.509 \\
Voxblox 6-thr &  &  &  & 1.333 \\ 
\midrule
\multicolumn{5}{l}{\textit{Dust data set}} \\
\midrule
OHM CPU
& 0.277 & 0.190 & 0.167 & 0.206 \\ 
OHM CUDA
& 3.064 & 1.963 & 1.885 & 2.945 \\ 
OHM OpenCL
& 1.645 & 1.069 & 0.903 & 1.433 \\ 
Octomap & 0.241 &  &  &  \\ 
Voxblox 1-thr & 0.239
&  &  & 0.701 \\
Voxblox 6-thr &  &  &  & 1.672 \\ 
\midrule
\multicolumn{5}{l}{\textit{Flatpack data set}} \\ 
\midrule
OHM CPU
& 0.344 & 0.226 & 0.202 & 0.242 \\
OHM CUDA
& 2.917 & 1.955 & 1.840 & 2.753 \\
OHM OpenCL
& 1.603 & 1.085 & 0.958 & 1.206 \\
Octomap & 0.241 &  &  &  \\
Voxblox 1-thr & 0.258
&  &  & 0.704 \\
Voxblox 6-thr &  &  &  & 1.933 \\
\midrule
\multicolumn{5}{l}{\textit{Hovermap data set}} \\
\midrule
OHM CPU
& 0.075 & 0.048 & 0.042 & 0.052 \\
OHM CUDA
& 0.898 & 0.714 & 0.599 & 0.851 \\
OHM OpenCL
& 0.379 & 0.268 & 0.229 & 0.338 \\
Octomap & 0.031 &  &  &  \\
Voxblox 1-thr & 0.019
&  &  & 0.086 \\
Voxblox 6-thr &  &  &  & 0.300 \\
\midrule
\multicolumn{5}{l}{\textit{Railway platform data set}} \\
\midrule
OHM CPU
& 0.367 & 0.272 & 0.239 & 0.269 \\
OHM CUDA
& 3.441 & 2.295 & 2.179 & 3.191 \\
OHM OpenCL
& 1.893 & 1.306 & 1.130 & 1.276 \\
Octomap & 0.320 &  &  &  \\
Voxblox 1-thr & 0.431
&  &  & 1.082 \\
Voxblox 6-thr &  &  &  & 2.300 \\
\bottomrule
\end{tabular}
\label{tab:ResOfflinePerformance}
\vspace{-0.5cm}
\end{table}

The number of rays per second able to be processed using the various methods is shown in Table~\ref{tab:ResOfflinePerformance}. The occupancy column includes calculation of mean for each voxel for OHM-based methods (but not Octomap or Voxblox, which do not maintain a voxel mean). OHM CUDA results utilise the laptop's discrete NVIDIA GPU, whereas OHM OpenCL results utilise integrated Intel graphics; both split rays into 10m segments, which was empirically found to perform best. OHM CPU results utilise a single thread. All methods truncate rays to a maximum range of 20m, based on the lidar's useful local mapping distance, and a voxel size of 10cm.

The GPU-based methods can be seen to provide 8-15$\times$ performance improvements in comparison to existing occupancy approaches found in Octomap and Voxblox. This large performance improvement enables processing of sensors with much higher data rates (as motivated in Table~\ref{tab:LidarSamplesPerSecond}), and multiple lidar units. It also enables real-time use of more methods requiring more computation, such as NDT-OM and NDT-TM. Although NDT-OM is around 30\% slower than the occupancy methods and NDT-TM is a further 5-15\% slower, both are well within real-time operation for the GPU-based implementations making them suitable for our robotic navigation applications.

The biggest difference among the data sets is for the Hovermap~\cite{hovermap} scene. As illustrated in Figure~\ref{fig:ResHovermap}, this is collected using a UAV mapping a communications tower at the peak of a hill. This very open scene has a large number of maximum-length rays, which necessitates update of the largest possible number of voxels. The higher performance in the Railway scene in Figure~\ref{fig:ResPlatform} corresponds to a more confined space. The Dust data set in Figure~\ref{fig:ResDust} is similarly confined, but significant dust produces more false positives.

\begin{figure}
	\vspace*{4pt}
	\centering
	\includegraphics[width=0.47\textwidth]{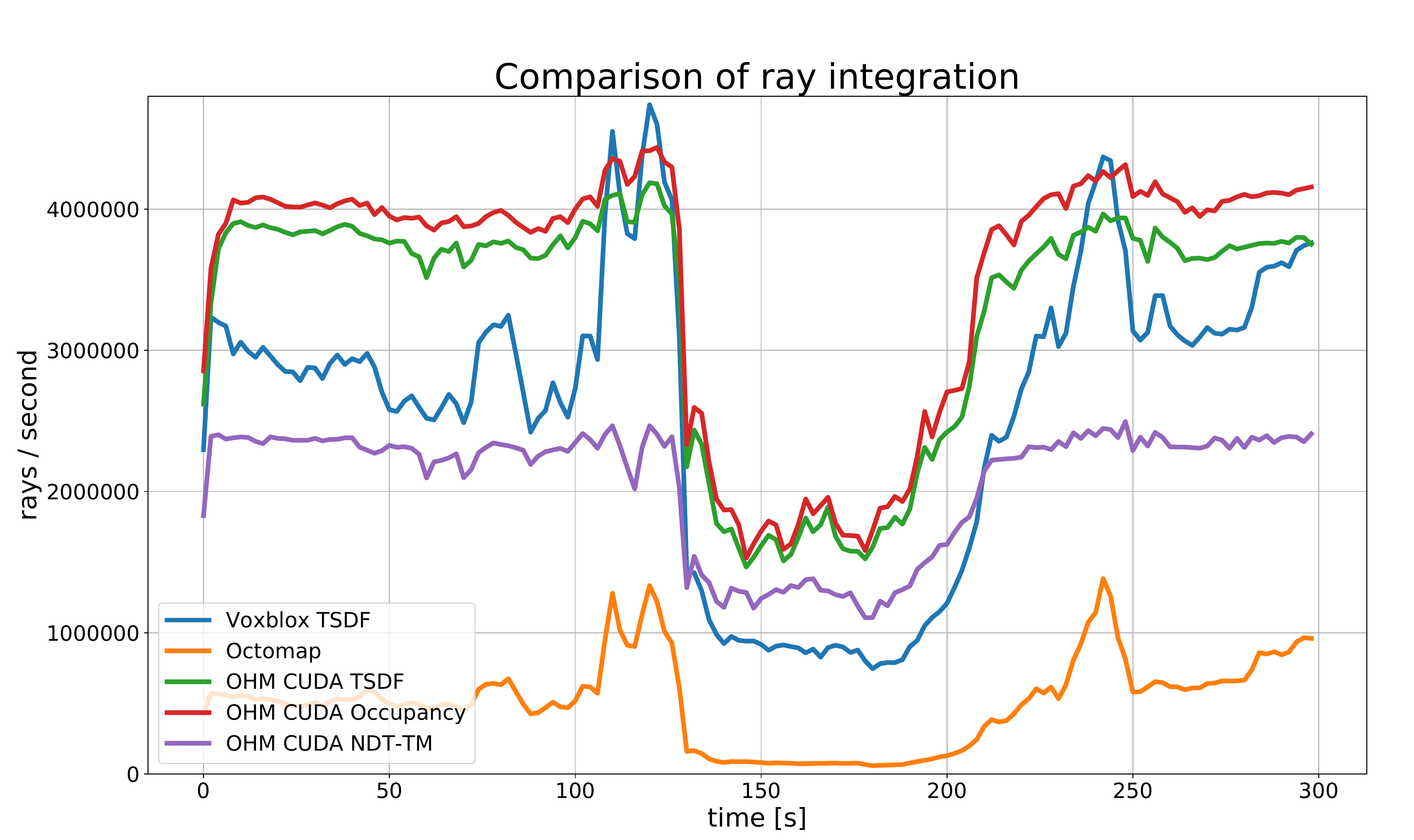}
	\caption{Rays/s achieved in offline processing by different methods on the Flatpack data set. Large changes correspond to transitions between indoors (with mostly short rays) and outdoors (with many maximum-length rays). The time axis shows data/ROS time rather than process time.}
	\label{fig:ResOfflineRaysPerSecondFlatpack}
	\vspace*{-14pt}
\end{figure}

The Flatpack data set incorporates both indoor and outdoor sections, which creates large changes in the computational load over time. This is examined in Figure~\ref{fig:ResOfflineRaysPerSecondFlatpack}. All methods show variation between indoors and outdoors. Voxblox shows more variation, understood to be during periods where the data-dependent optimisation becomes more or less effective.

The Voxblox TSDF results show excellent performance compared to the CPU-based occupancy methods. The results are shown for one and six threads using \emph{fast} processing, which traces rays backwards and terminates processing upon reaching a voxel that has already been updated twice within the scan. This optimisation is not utilised in the OHM GPU implementation of TSDF, and is somewhat unsuited to GPU operation, where all threads in a thread group need to execute the same instructions. The OHM TSDF implementation is considered experimental, and is provided for timing comparison. In all cases, the CUDA GPU performance is 50-100\% better than the six thread Voxblox fast processing.

Even if processing of high-rate data is possible in both CPU and GPU implementations, in many real-time embedded systems, CPU resources are in higher demand by processes that cannot be easily parallelised such as motion planning and control. Hence, the ability to offload processing of high-rate data to the GPU is highly beneficial.

\subsection{Online Results}
\begin{figure}
	\vspace*{4pt}
	\centering
	\includegraphics[width=0.48\textwidth]{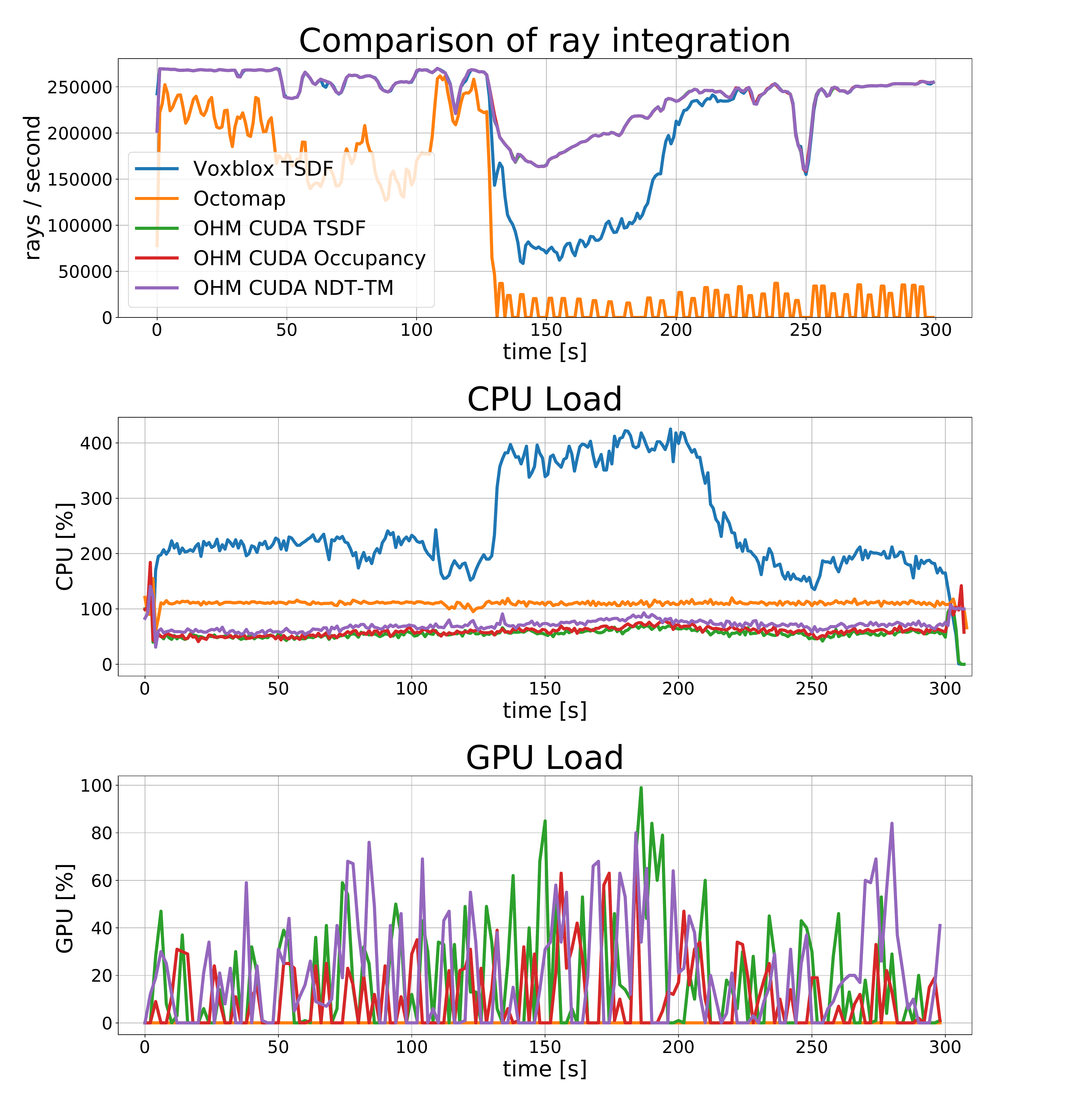}
	\caption{Online operation for the Flatpack data set. Top figure shows rays/s processed (indicating if method kept up with the data rate), middle shows CPU consumption, and bottom shows the GPU resource consumption. The OHM CUDA TSDF line in the top plot is essentially coincident with the Occupancy and NDT-TM lines.}
	\label{fig:ResOnlineXavier}
	\vspace*{-14pt}
\end{figure}
Online test results are based on both NVIDIA Jetson Xavier (CUDA) and Intel NUC8i7BEH NUC (Integrated Intel Iris Plus Graphics 655 through OpenCL 1.2) platforms. We focus analysis on the Flatpack data set shown in Figures~\ref{fig:ResFlatpack} and \ref{fig:ResOfflineRaysPerSecondFlatpack} since it contains segments of both enclosed and outdoor environments. In the top plots in Figure~\ref{fig:ResOnlineXavier}, we show the CPU and GPU utilisation of the various approaches in the data set. We also show the rays processed per second, which indicates whether the algorithm is able to keep up with the data, or needs to drop rays. The top graph shows that real-time operation is enabled by the GPU-based methods in both indoor and outdoor environments, while Voxblox only achieves real-time performance in the indoor environments with shorter rays (shown by the drop in the number of rays processed in the time 130-180\,s). Octomap is unable to process the data during the beginning indoor period, and suffers a large performance degradation when the outdoor area expands the map, causing deeper traversal of the octree in voxel queries. Voxblox's \textit{fast} TSDF processing keeps up indoors on the Xavier using two full cores (206\% CPU usage), but drops 31\% of rays outdoors even while occupying four cores (378\%). On the NUC, it keeps up indoors with 246\% CPU usage, and drops around 2\% of rays outdoors with 334\% CPU usage.

For occupancy, OHM utilises an average 9\% of GPU resources indoors and 19\% outdoors on the Xavier, and 8\% and 16\% respectively on the NUC (plots for the NUC are not shown due to space limitations). For NDT-TM, these figures are 19\% and 22\% respectively on the Xavier and 18\% and 31\% on the NUC. The CPU resources occupied in these cases are around 55\% and 66\% on the Xavier, and 58\% and 66\% on the NUC (for occupancy; NDT-TM is around 11\% higher on the Xavier and 2\% higher on the NUC). A negligible number of rays go unprocessed in all OHM cases.

\subsection{Deployment in the DARPA Subterranean Challenge}
\label{ss:SubT}
OHM forms a key part of the local navigation solution used by the Team CSIRO Data61 in SubT~\cite{HudTal21,hines2021}, where it has been used in a variety environments including mine tunnels, an unfinished nuclear power plant and natural caves. OHM was utilised on all ground-based vehicles, including large and small tracked vehicles, multiple quadrupeds, and a large legged hexapod. All vehicles were equipped with the same perception payload with a spinning Velodyne VLP-16 Puck lidar as a key component. On the Spot quadruped, the integrated depth cameras were also incorporated in OHM to complement the lidar information in the near range. The camera operated at 5\,Hz with its range limited to 2\,m and $6\times$ down-sampling resolution.

The OHM map was set as a robot-centric solution fixed to a 20\,m$\times$20\,m$\times$20\,m volume using 0.1\,m voxel resolution in all platforms. Its map was the precursor used to generate a height map for local path planning generated at approximately 5\,Hz. OHM was run on an Intel NUC8i7BEH with integrated graphics on the smaller platforms, and a Cincoze DX-1100 using an Intel i7-8700T on the larger tracked platforms, in both cases shared with other navigation CPU based processes.

OHM was a key enabler our team's solution, which placed second at the Final Event of SubT. The 3D occupancy maps enabled the development of virtual surfaces as a fundamental approach for handling negative obstacles~\cite{hines2021}. The integration of lidar and depth camera data on the Spot platform provided coverage at steep elevation angles which resulted in successful autonomous ascent and descent of stairs in the final event. The use of the GPU to process these streams freed the CPU for the rest of the navigation stack, and avoided the need for additional or higher power computational platforms. The light weight of the payload is understood to be a significant factor in the stability of the platform on challenging surfaces, and thus OHM contributed the strong outcome in many ways.

\section{Conclusion}
\label{sec:Conclusion}
We have presented an open source, GPU-based OGM algorithm that enables use of high data rate sensors, and is suitable for online and offline processing. We have shown how this algorithm provides significant performance improvements over state-of-the-art algorithms though analysis of real-world data sets on embedded platforms. Finally, we have shown how the flexible architecture accommodates a range of modern extensions, providing a solid basis for further development.

\addtolength{\textheight}{-0cm}

\bibliographystyle{IEEEtran}
\bibliography{ohm}

\end{document}